\documentclass{MITcsail}

\usepackage{amsmath,amsfonts,bm}

\def\eqref#1{equation~\ref{#1}}

\def\1{\bm{1}}

\DeclareMathAlphabet{\mathsfit}{\encodingdefault}{\sfdefault}{m}{sl}
\SetMathAlphabet{\mathsfit}{bold}{\encodingdefault}{\sfdefault}{bx}{n}

\usepackage{graphicx} 
\usepackage{hyperref}
\usepackage{amsmath, amssymb, amsfonts} 
\usepackage{xurl}
\usepackage{subcaption}
\usepackage{tcolorbox}
\usepackage{amsmath, amssymb, amsfonts}
\usepackage{xurl} 
\usepackage{booktabs} 

\usepackage{makecell} 

\tcbuselibrary{breakable}
\newtcolorbox{promptbox}[2][]{
    colback=gray!5!white,
    colframe=gray!75!black,
    breakable,
    title=#2,
    #1
}

\title{Population-Evolve: a Parallel Sampling and Evolutionary Method for LLM Math Reasoning}
\author
{Yanzhi Zhang~$^{1,2,3}$, 
\textbf{Yitong Duan~$^{3,4}$\footnote{Correspondence. E-mail: \texttt{duanyitong@zgci.ac.cn}}},
Zhaoxi Zhang~$^{4,5}$,
\textbf{Jiyan He~$^{3,4}$},
\textbf{Shuxin Zheng~$^{3,4}$}
\\
\vspace{1em}
\normalfont{\small $^{1}$Academy of Mathematics and Systems Science, Chinese Academy of Sciences}\\
\normalfont{\small $^{2}$University of Chinese Academy of Sciences}\\
\normalfont{\small $^{3}$Zhongguancun Academy}\\
\normalfont{\small $^{4}$Zhongguancun Institute of Artificial Intelligence} \\
\normalfont{\small $^{5}$Peking University}\\
\vspace{2em}
}

\begin{document}

\maketitle

\begin{abstract}
    Test-time scaling has emerged as a promising direction for enhancing the reasoning capabilities of Large Language Models in last few years. In this work, we propose Population-Evolve, a training-free method inspired by Genetic Algorithms to optimize LLM reasoning. Our approach maintains a dynamic population of candidate solutions for each problem via parallel reasoning. By incorporating an evolve prompt, the LLM self-evolves its population in all iterations. Upon convergence, the final answer is derived via majority voting. Furthermore, we establish a unification framework that interprets existing test-time scaling strategies through the lens of genetic algorithms. Empirical results demonstrate that Population-Evolve achieves superior accuracy with low performance variance and computational efficiency. Our findings highlight the potential of evolutionary strategies to unlock the reasoning power of LLMs during inference.
\end{abstract}

\section{Introduction}

In recent years, the advent of Large Language Models (LLMs) has profoundly advanced the frontier of artificial general intelligence. A critical benchmark for this progress is the ability of models to perform complex mathematical reasoning. As LLMs have demonstrated rapidly advancing capabilities, numerous mathematical benchmarks have been developed to evaluate their reasoning abilities. The field has witnessed a rapid escalation in the difficulty of solvable problems, progressing from grade-school level math questions like \texttt{GSM8K}~\citep{cobbe2021gsm8k} to challenging high-school competition math problems such as \texttt{AMC} and \texttt{AIME}. Most recently, Google DeepMind~\citep{deepmind2025deepthink}, OpenAI~\citep{openai2025imo} and DeepSeek~\citep{deepseek-math-v2} all achieved gold-medal–level performance at IMO 2025 by solving five out of six problems. The success on these challenging competition math frontiers is particularly noteworthy, as it demonstrates human-like deep reasoning abilities considered fundamental to the pursuit of Artificial General Intelligence.


To solve complex math problems, researchers have been trying to improve LLM's reasoning abilities. An early and notable example of this effort is the Chain-of-Thought (CoT)~\citep{wei2022} technique, which enhances the mathematical reasoning of LLMs through prompting.
Then researchers have employed supervised fine-tuning (SFT)~\citep{wei2022,lewkowycz2022solvingquantitativereasoningproblems} and reinforcement learning with human feedback (RLHF)~\citep{ouyang2022traininglanguagemodelsfollow,lightman2023letsverifystepstep,wang2024mathshepherdverifyreinforcellms} to increase the mathematical reasoning ability of LLMs. A significant breakthrough was achieved with DeepSeek-R1~\citep{deepseekai2025deepseekr1incentivizingreasoningcapability}, which applied reinforcement learning with verifiable rewards (RLVR) to enhance model reasoning. This method incentivized the model to internalize the step-by-step process of COT, effectively transforming it from a prompted technique into an inherent capability. Currently, the community is using model distillation to generate mathematical question-answering pairs with reasoning process from a teacher model. These datasets are then used to SFT smaller LLMs, followed by RLVR, to enhance their reasoning abilities~\citep{moshkov2025aimo2winningsolutionbuilding}. 

However, these post-training methods are hindered by expensive computation and time costs, and the scarcity of high-quality data for post-training, leading to diminishing marginal returns.

As the capabilities of state-of-the-art open-source models have grown, training-free paradigm to improve math-reasoning has emerged within the research community.  Training-free methods bypass these issues by leveraging a powerful LLM's innate reasoning abilities to improve its own performance on a given task. They require no additional post-training data collection, impose minimal computational resources, and can be applied instantly at inference, thus making them both cost‑efficient and time‑efficient. These training-free methods can be broadly categorized into the following three types: (1) Selecting after sampling~\citep{wang2023selfconsistencyimproveschainthought,toshniwal2025genselectgenerativeapproachbestofn} (2) Sequential evolving~\citep{imo_gemini_2.5_pro,dser,know_flow}.



While GenSelect in \cite{toshniwal2025genselectgenerativeapproachbestofn} uses LLM's verifying and selecting better solutions abilities to enhance model's performance, methods like DSER~\citep{imo_gemini_2.5_pro,dser} and Knowledge-Flow~\citep{know_flow} have shown that iterative refinement can significantly boost performance on complex math problems. However, such iterative refinement methods suffer from high variance of performance and high time cost due to the ``serial iteration". At the same time, GenSelect does not iterate continuously, thus failing to fully leverage the generative and verification capabilities of the large model.

To address this challenge—namely, the high performance variance and substantial time cost introduced by serial iterative refinement in DSER, as well as GenSelect’s inability to fully exploit continuous generative‑verification cycles—we propose a new method called \textit{Population-Evolve}. The name is inspired by population evolution in Biology: we iteratively `evolve' a population of $P$ candidate solutions in each iteration, guiding the model to generate progressively better results.  As the LLM produces the next generation of candidates, it implicitly compares all individuals from the previous generation, yielding a new population with higher average quality. Moreover, as the number of iterations increases, the population gradually converges, resulting in a highly stable method.

In summary, the core contributions of this work are as follows.
\begin{enumerate}
    \item We propose a novel parallel sampling and evolutionary method, named \textit{Population-Evolve}, for LLM math reasoning.
    \item We present a unified framework for training-free paradigms that improve LLM reasoning from the perspective of genetic algorithms.
    \item Our experimental results demonstrate the significant potential of test-time-scaling in LLM mathematical reasoning tasks.
\end{enumerate}

\begin{figure}[H]
\centering
\includegraphics[width=1.0\textwidth]{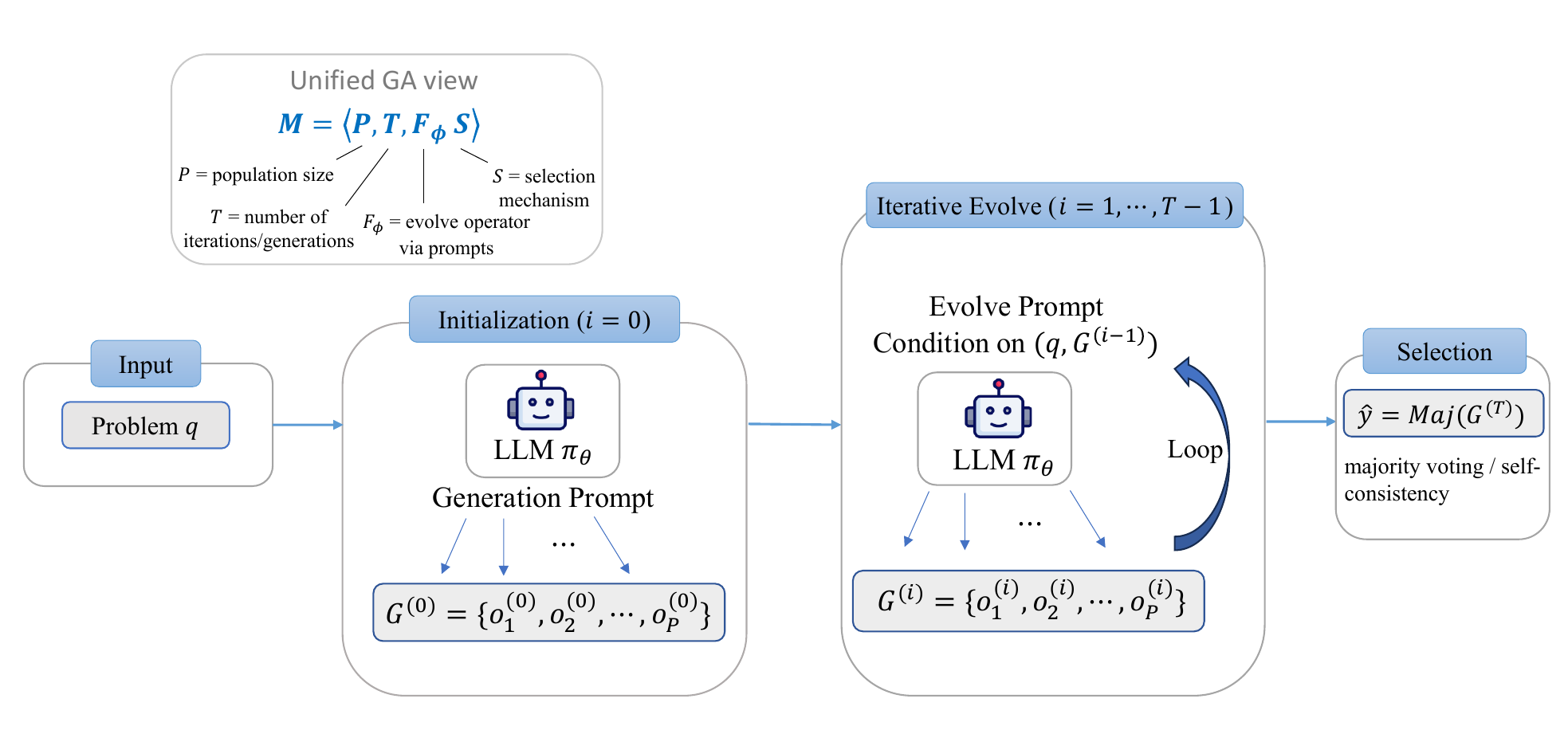}
\caption{\textbf{An Overview of the Proposed Population-Evolve.}}
\label{fig:whole}
\end{figure}

\section{Related works}
In this section, we will discuss several important research works related to our topics.

\subsection{Math Reasoning}
\subsubsection*{Natural‑Language vs. Formal‑Language Mathematical Reasoning}
Research on LLMs for mathematical reasoning has diverged into two closely related but distinct paths-natural language and formal theorem proving. In the natural‑language setting, general‑purpose and math‑specialized models—such as GPT‑class models, DeepSeek‑R1 and DeepSeek-Math‑V2~\citep{deepseek-math-v2}—are typically evaluated on math competition problems by generating natural‑language solutions. In parallel, some work targets formal theorem proving, where systems built on proof assistants such as Lean, Coq, and Isabelle—including large formal provers like DeepSeek‑Prover‑V2~\citep{ren2025deepseekproverv2advancingformalmathematical} for Lean 4—translate mathematical statements into fully formal languages and use LLMs to propose proof sketches that are subsequently checked by a mechanized verifier.


\subsubsection*{Math Reasoning Systems}
Several AI systems extend the landscape of AI mathematical reasoning. AlphaProof~\citep{alphaproof} combines Gemini‑class language models with search and reinforcement‑learning techniques to translate natural‑language mathematical problems into Lean and explore proof spaces that can be validated by a formal proof checker. AlphaGeometry~\citep{alphageometry} and its successor AlphaGeometry2~\citep{chervonyi2025goldmedalistperformancesolvingolympiad} focus specifically on geometric problem solving, using symbolic numeric pipelines to generate solutions to competition‑math geometry problems. Gemini-DeepThink is an advanced reasoning mode for the Gemini family of models by using parallel thinking. It was officially graded and certified by the IMO committee, scoring 35 out of 42 points, thus matching the gold threshold~\citep{deepmind2025deepthink}

\subsection{Training-free Methods for LLM Math Reasoning}
In this part, we discuss some of the popular training-free work that enhance LLM's reasoning performance in the context of the math reasoning task. 

\subsubsection*{Sequential Evolving}

Literature on LLM performance through sequential iteration focuses on both inference-time refinement and training methods. For instance, \cite{madaan2023selfrefineiterativerefinementselffeedback} demonstrates that LLM can enhance their performance by iteratively generating internal feedback and refining their outputs themselves without any external feedback across diverse tasks. 
\cite{zhao2025boostingllmreasoningspontaneous} introduce an approach called SPOC, a spontaneous self-correction framework that leverages synthetic data fine-tuning to internalize a dual-role proposer-verifier dynamic within LLM on math reasoning, enabling them to generate solutions and verifications in a single pass that dynamically terminate to effectively scale inference-time compute.

One notable study by~\cite{imo_gemini_2.5_pro} introduces a ``verification‑and‑correction" pipeline, applied to IMO 2025 problems. Using leading models such as Gemini-2.5-Pro, Grok‑4, and GPT‑5, the pipeline solved five out of six problems, a substantial gain over baseline performances of these models. 

A natural question arises: can such a self-evolving loop be applied to smaller open-source language models? To our best knowledge, \cite{dser} and \cite{know_flow} are the first two studies to explore this direction. The \path{DSER} paper tests the framework with simpler prompts and a streamlined process, evaluating their method on the \path{DeepSeek-Distill-Qwen3-8B} model using the \path{AIME’25} and \path{AIME’24} datasets. They found that after long serial self-evolving iterations, the final accuracy surpass the \path{avg@16} metric of the teacher model-\texttt{DeepSeek-R1-2508}.

The \path{Knowledge-Flow} study systematically open-sourced their implementation and evaluated it on various models, including \path{gpt-oss} series models, \path{Qwen3} series models. They found that their approach significantly enhances the performance of larger models such as \path{gpt-oss-120B}~\citep{openai2025gptoss120bgptoss20bmodel} and \path{Qwen3-235B-A30B}~\citep{yang2025qwen3technicalreport}, allowing them to achieve 100\% accuracy on the \path{AIME’25} dataset and surpass their \path{maj@64} baseline. However, their method fails on smaller models like \path{gpt-oss-20B}~\citep{openai2025gptoss120bgptoss20bmodel}, \path{Qwen3-4B-Thinking}~\citep{yang2025qwen3technicalreport}, and \path{Qwen3-30B-A3B-Thinking}~\citep{yang2025qwen3technicalreport}, as reflected by the fact that the final accuracy does not surpass the \path{maj@64} score.


\subsubsection*{Selecting After Parallel Sampling}

The study by \cite{snell2024scalingllmtesttimecompute} highlights the potential performance gains of test-time compute. Furthermore, \citet{wang2023selfconsistencyimproveschainthought} demonstrates the significant performance improvements achieved by simply applying majority voting to the answers derived from parallel reasoning. After parallel sampling, which produces multiple independent candidate solutions, the central and differentiating task is to identify the best response from the pool of candidates. Approaches for selecting solution include rule-based methods~\citep{hassid2025dontoverthinkitpreferring}, discriminative reward models~\citep{liu2025acemathadvancingfrontiermath}, and the recent GenRM~\citep{zhang2025generativeverifiersrewardmodeling,shi2025heimdalltesttimescalinggenerative}, as well as GenSelect~\citep{toshniwal2025genselectgenerativeapproachbestofn}, which currently achieves state-of-the-art performance.


The methods we mentioned above have been thoroughly discussed in paper \cite{toshniwal2025genselectgenerativeapproachbestofn}. They found out the GenSelect (generative select) better than GenRM (generative reward model). This suggests that large models are more adept at making comparative judgments to identify the better response than at assessing the absolute quality of a single response. The study's core contribution is the introduction of GenSelect, a method that is shown to significantly outperform prior approaches on various competition-level math reasoning benchmarks. The authors also report that the method's performance is stable across different inference configurations, allowing for highly efficient scaling. 


\section{Preliminaries}
In this section, we provide a brief overview of our proposed method and relevant baselines. 

We introduce an iterative framework for solution generation and refinement. 
Given prefix context $x$, we denote the distribution of LLM with parameter $\theta$ is $\pi_{\theta}(\cdot | x)$.

\subsection{Unified Framework for Solution Generation and Refinement}

We unify our proposed method and the relevant methods into a general framework, inspired by Genetic Algorithms and AlphaEvolve~\citep{novikov2025alphaevolvecodingagentscientific}. We define the whole generation process as a tuple $\mathcal{M} = \langle P, T, \mathcal{F}_{\phi}, \mathcal{S} \rangle$, where $P$ denotes the population size (parallelism), $T$ denotes the maximum number of evolution iterations (or called maximum generations in GA), $\mathcal{F}_{\phi}$ represents the evolutionary operators conditioned on a prompt set $\phi$, and $\mathcal{S}$ is the final selection mechanism. The fitness function that evaluate the quality of individuals, is then the reward function that verify whether the solution of given math question is correct. We use $G^{(i)}$ to denote the $i$-th solutions in iteration, or called the $i$-th generation in GA.

Given a query $q$ from a dataset $\mathcal{D}$, the generalized process proceeds in three stages:

\paragraph{- Initialization (Population Sampling).}
Regardless of the method, the process begins by sampling an initial population of candidate solutions $G^{(0)} = \{o_1^{(0)}, \dots, o_P^{(0)}\}$ from the model distribution:
\begin{equation}
    o_j^{(0)} \sim \pi_{\theta}(\cdot | q), \quad j \in \{1, \dots, P\}
\end{equation}

\paragraph{- Iterative Evolution.}
For iterations $i = 0$ to $T-1$, the set of solutions $G^{(i)}$ is mapped to $G^{(i+1)}$ via the evolutionary operators $\mathcal{F}_{\phi}$. This function uses the prompt utilized to refine or verify solutions to get new generations:
\begin{equation}
    G^{(i+1)} \leftarrow \mathcal{F}_{\phi}(G^{(i)}, q; \pi_{\theta})
\end{equation}
If $T=0$, this stage is skipped (static generation).

\paragraph{- Solution Selection.}
After $T$ iterations, a final selection mechanism operator $\mathcal{S}$ aggregates or analyzes the final population $G^{(T)}$ (and potentially interaction history) to output the final answer $\hat{o}$:
\begin{equation}
    \hat{o} = \mathcal{S}(G^{(T)} | q)
\end{equation}

\noindent Below, we instantiate the three distinct test-time-scaling methods within this unified framework:

\paragraph{Population-Evolve (Our Method).}
This method combines parallel sampling with iterative refinement using a unified evolving prompt.
\begin{itemize}
    \item \textbf{Parallelism:} $P > 1$.
    \item \textbf{Evolution ($T \ge 1$):} The evolution function applies a uniform ``evolving prompt'' $p_e$. Each solution evolves independently:
    \begin{equation}
     o_j^{(i+1)} \sim \pi_{\theta}(\cdot | q, p_e, G^{(i)}) \quad \text{for}\ 1\leq j\leq P
    \end{equation}
    \item \textbf{Selection:} We obtain $P$ solutions following the evolution process. As the number of iterations increases, the answers tend to converge. Therefore, we typically employ Majority Voting (Self-Consistency)~\citep{wang2023selfconsistencyimproveschainthought} over the final set $G^{(T)}$ to get every final answer for the dataset.
\end{itemize}

\paragraph{Deep Self-Evolving Reasoning (DSER).}
Proposed by \cite{imo_gemini_2.5_pro,dser}, this method focuses on the deep refinement of a single trajectory through a verify-and-refine cycle.
\begin{itemize}
    \item \textbf{Parallelism:} $P = 1$ (Single trajectory).
    \item \textbf{Evolution ($T \ge 1$):} The evolution function is composite, consisting of a verification step (with prompt $p_v$) and a refinement step (with prompt $p_r$). For the single candidate $o^{(i)}$:
    \begin{align}
        v^{(i)} &\sim \pi_{\theta}(\cdot | q, o^{(i)}, p_v) \nonumber \\
        o^{(i+1)} &\sim \pi_{\theta}(\cdot | q, o^{(i)}, p_v, v^{(i)}, p_r)
    \end{align}
    \item \textbf{Selection:} The final refined output $o^{(T)}$ is adopted directly. 
\end{itemize}

\paragraph{GenSelect.}
Proposed by \cite{moshkov2025aimo2winningsolutionbuilding}, this method focuses on discriminator-based selection rather than iterative content modification.
\begin{itemize}
    \item \textbf{Parallelism:} $P > 1$.
    \item \textbf{Evolution ($T = 0$):} No iterative refinement is performed on the solutions. So we get $G^{(0)}={o_1^{(0)},o_2^{(0)},\cdots,o_P^{(0)}}$.
    \item \textbf{Selection:} The method employs the LLM itself as a discriminator using a specific selection prompt $p_s$. The model analyzes the entire context of candidates to choose the index of the best solution:
    \begin{equation}
        j^* \sim \pi_{\theta}(\cdot | q, G^{(0)}, p_s), \quad \hat{o} = o_{j^*}^{(0)}
    \end{equation}
    Following the evaluation in \cite{toshniwal2025genselectgenerativeapproachbestofn}, we also perform Majority Voting~\citep{wang2023selfconsistencyimproveschainthought} over the answers selected by GenSelect.
\end{itemize}

\subsection{Naïve Baselines}

\paragraph*{Pass@N.} This is an oracle baseline, which serves as the upper bound for LLM's performance. We use the unbiased estimator in \cite{chen2021evaluatinglargelanguagemodels} to calculate this metric.
\begin{equation}
\label{eq:pass}
\textbf{pass@k}=\mathbb{E}_{\text{problems}}[1-\frac{\binom{n-c}{k}}{\binom{n}{k}}]
\end{equation}

\paragraph*{Majority Voting/Self-Consistency.} Proposed by \cite{wang2023selfconsistencyimproveschainthought}, this method selects the most common answer 
from the solution candidates.
\begin{equation}
\label{eq:maj}
\text{Majority}(\{\text{Ans}(o_1), \dots, \text{Ans}(o_P)\})
\end{equation}

\section{Experiment}

\subsection{Evaluation Details}
The AIME, BRUMO and HMMT competitions are popular and challenging benchmarks for math reasoning. 
We compile the HMMT24-25 dataset by combining MathArena's \verb|hmmt_feb_2024| and \verb|hmmt_feb_2025|~\citep{balunovic_srimatharena_2025} from huggingface, for a total of 60 challenging problems. We also collect AIME25~\citep{aime25} dataset and BRUMO25~\citep{balunovic_srimatharena_2025}, which contain 30 problems each.

Our experiments are mainly conducted on three open-source LLMs: \path{Qwen3-4B-Thinking-2507}, \path{DeepSeek-R1-0528-Distill-Qwen3-8B}, and \path{OpenReasoning-Nemotron-7B}. For all trials, we set the temperature to 0.6 to balance the diversity of rollouts and their accuracy. To accurately parse and extract answers from the solutions, we employed the Math-Verify library~\citep{Kydlicek_Math-Verify_Math_Verification}.

\subsection{Prompts Used in Experiments}
\subsubsection{Population-Evolve Prompts} \label{para:prompts}

To implement our Population-Evolve methodology, we designed a two-stage prompting strategy. This strategy is comprised of a Generation Prompt for the initial step and an Evolving Prompt for the core evolutionary loop.

The Generation Prompt, shown below, is a straightforward CoT prompt. Our purpose is to obtain a set of initial reasoning paths from LLM for a given mathematical problem, by generating an initial population of $P$ solutions, denoted as $G^{(0)} = \{o_1^{(0)}, o_2^{(0)}, ..., o_P^{(0)}\}$. 

\begin{promptbox}{Generation Prompt}
Please think step by step and put your final answer in \verb|\\boxed{{}}|.

Question:
\verb|{question}|
\end{promptbox}

The core prompt of our method is the Evolving Prompt. After the initial generation, for each subsequent iteration $i \ge 0$, we employ this prompt to guide the LLM toward more accurate and robust solutions. As detailed below, the prompt provides the LLM with the original problem alongside the entire population of solutions, $G^{(i)}$, from the preceding iteration.

Following standard processing procedure for reasoning models, we preprocess the strings $o_{j}^{(i)}$ in $G^{(i)}$ by removing the part within \verb|<think>...|\verb|</think>| tags and retaining the final response, which has already illustrated the detailed solution process.

This preprocessing is crucial for two reasons. First, the content within the \verb|<think>...| \verb|</think>| is excessively long; removing it significantly reduces the context length, allowing more samples to be processed. Second, since the final solution already encapsulates the results of the reasoning process, this step effectively reduces noise without sacrificing essential information.
\begin{promptbox}{Evolving Prompt}
You will be given a challenging math problem followed by \verb|{len(solutions)}| solutions. The solutions can be wrong or right. Your task is to understand such solutions and then generate a detailed new solution.\\

Problem: \verb|{Question Text}|\\

Solutions:

--- Solution 0 ---

\verb|{Solution Text 0}|

--- Solution 1 ---

\verb|{Solution Text 1}|

...\\

Your final response should include: \\
- Detailed new solution.
\end{promptbox}

\subsubsection{Prompts in Other Methods}

The prompts used in DSER and GenSelect can be seen in Appendix~\ref{ax:prompts}.

\subsection{Main Results}
In Table~\ref{tab:main_results}, we compare three open‑source reasoning LLMs on the HMMT24–25 and AIME25 datasets using three different test‑time scaling methods: DSER, GenSelect, and our proposed Population‑Evolve approach.

Our method uses $P=16$, max iterations $T=16$ (most of time, it converges before 16, but we use 16 for simplicity).  \textbf{To ensure robustness,  four independent trails were conducted, and the mean accuracy avg@4 is reported.}  The prompts can be seen in section~\ref{para:prompts}.

For DSER, we set the maximum number of iterations to $T = 32$ (most of time, it converges between 25 to 32 turns).  To reduce the high variance typically observed with this method, we perform $16$ parallel runs and report results in the format avg@16, where avg@16 is the average performance across runs. The prompts can be seen in~\ref{ax:prompts}.

For GenSelect, we prompt the LLM to generate $16$ candidate solutions per question, repeat the selection process $32$ times, and then apply majority voting over the selected solutions to obtain the final answer. We adopt this configuration as it aligns with the final settings reported in \cite{moshkov2025aimo2winningsolutionbuilding} to \textbf{maximize the method's performance.} The prompts can be seen in~\ref{ax:prompts}. \textbf{To ensure robustness,  four independent trails were conducted, and the mean accuracy avg@4 is reported.}

To compare with naïve baselines, we also compute maj@k using majority voting over 64 independent rollouts per question. Surprisingly, all three test-time scaling methods outperform this strong baseline. As an oracle baseline, we include pass@16 (computed using Equation~\ref{eq:pass}), which represents an upper bound determined by the reasoning limitations of the LLM. 



On the HMMT24–25 dataset, our proposed Population‑Evolve method achieves 64.17\% accuracy on \path{Qwen3-4B-Thinking-2507}, substantially outperforming Maj@64 (58.33\%), and the DSER baseline 61.46\% and GenSelect baseline 63.75\%. For the \path{OpenReasoning‑Nemotron‑7B} model, our method reaches 82.08\%, exceeding DSER baseline 77.08\% and Genselect 78.75\%. 
Likewise, on \path{DeepSeek-R1-0528-Distill-Qwen3-8B}, our approach attains 70.83\%, surpassing Maj@64 (65.00\%), DSER (65.31\%) and GenSelect (68.75\%).

On the comparatively easier AIME25 dataset, our method also achieves the best performance across all three models, reaching 88.33\% accuracy on \path{DeepSeek-R1-0528-Distill-Qwen3-8B} and 90.00\% on \path{Qwen3-4B-Thinking-2507}, and further attaining 93.33\% on \path{OpenReasoning-Nemotron-7B}. 

On the BRUMO25 dataset, considered the least challenging, our method still ranked first with 93.33\% accuracy on both \path{Qwen3-4B-Thinking-2507} and \path{DeepSeek-R1-0528-Distill-Qwen3-8B}. On the \path{OpenReasoning-Nemotron-7B} model, our method reaches 95.83\%.

This consistent improvement demonstrates that the Population‑Evolve strategy not only enhances robustness across challenging reasoning benchmarks but also maintains superior effectiveness as task difficulty decreases, underscoring its broad applicability and generalization capability. Notably, we observe that as problem complexity increases, the margin by which our method outperforms the maj@k baseline becomes increasingly pronounced.

To quantify this, we calculated the average performance increase of our propsed Population-Evolve over Maj@64 across the three models for each benchmark: on the most difficult benchmark, HMMT24-25, our method achieves an average accuracy improvement of 7.92\%. While on the moderately difficult AIME25, the average gain is 5.00\%. As for the least difficult benchmark, BRUMO25, the average gain is 4.17\%.

We also evaluated top-tier LLMs like Google's Gemini-2.5-Pro and OpenAI's GPT-5.1 by using the OpenRouter API. For their evaluation, we enabled the thinking mode and set the max generation length to $57,600$, consistent with our main experiments. Unexpectedly, their pass@1 score, calculated as the average of four rollouts, was lower than that of smaller open-source LLMs enhanced with test-time scaling methods. This reveals the significant potential of test-time-scaling in LLM math reasoning tasks.


\begin{table}[H]
    \centering
    \caption{\textbf{Performance Comparison on Mathematical Reasoning Benchmarks.} We compare our proposed method with DSER, GenSelect, and Majority Voting across three models. The best results are highlighted in \textbf{bold}, and the second-best are \underline{underlined}. (Note that pass@k serves as an oracle upper bound and is not included in the comparison.)}
    
    \label{tab:main_results}
    \vspace{0.2cm} 
    \resizebox{\columnwidth}{!}{ 
        \scriptsize 
        \renewcommand{\arraystretch}{0.9} 
        \setlength{\tabcolsep}{4pt} 
    \begin{tabular}{lcccc}
        \toprule
        \textbf{Method} & \textbf{HMMT24-25} & \textbf{AIME25} & \textbf{BRUMO25} & \textbf{Avg} \\
        \midrule
        
        \multicolumn{5}{c}{\textbf{Qwen3-4B-Thinking-2507}} \\
        \cmidrule(lr){1-5}
        pass@16(oracle) & 76.61 & 91.45 & 96.41 & 88.16 \\
        pass@1(avg@64) & 53.15 & 80.78 & 80.52 & 71.48 \\
        Maj@64 & 58.33 & 86.67 & 86.67 & 77.22 \\
        DSER & 61.46 & 85.42 & 85.62 & 77.50 \\ 
        GenSelect & \underline{63.75} &  \underline{89.17} & \underline{88.33} &  \underline{80.42} \\ 
        \textbf{Our Method} & \textbf{64.17} & \textbf{90.00} & \textbf{93.33} &  \textbf{82.50}\\
        \midrule
        
        \multicolumn{5}{c}{\textbf{DeepSeek-R1-0528-Distill-Qwen3-8B}} \\
        \cmidrule(lr){1-5}
        pass@16(oracle) & 81.51 & 90.28 & 95.92 & 89.24 \\
        pass@1(avg@64) & 53.57 & 75.00 & 79.38 & 69.32 \\
        Maj@64 & 65.00 & 83.33 & \underline{90.00} & 79.44 \\
         DSER & 65.31 & \underline{85.42} & 88.96 & 79.90\\
        GenSelect & \underline{68.75} & 85.00 & \underline{90.00} & \underline{81.25} \\
        \textbf{Our Method} & \textbf{70.83} & \textbf{88.33} & \textbf{93.33} &  \textbf{84.16} \\
        \midrule
        
        \multicolumn{5}{c}{\textbf{OpenReasoning-Nemotron-7B}} \\
        \cmidrule(lr){1-5}
        pass@16(oracle) & 88.09 & 95.54 & 97.27 & 93.63 \\
        pass@1(avg@64) & 57.97 & 79.79 & 80.10 & 72.62 \\
        Maj@64 & 70.00 & 86.67 & 93.33 & 83.33 \\
        DSER & 77.08 & \underline{91.88} & \underline{94.79} & 87.92\\
        GenSelect & \underline{78.75} & 91.67 & 94.17 & \underline{88.20} \\
        \textbf{Our Method} & \textbf{82.08} & \textbf{93.33} & \textbf{95.83} & \textbf{90.41} \\
        
        \midrule[1pt] 
        \multicolumn{5}{c}{\textit{Top-tier LLMs}} \\
        \cmidrule(lr){1-5}
        \textbf{Gemini-2.5-pro} (pass@1(avg@4)) & 57.92 & 80.83 & 79.17 & 72.64 \\
        \textbf{gpt-5.1} (pass@1(avg@4)) & 72.08 & 86.67 & 86.67 & 81.81 \\
        \bottomrule
    \end{tabular}
    }
\end{table}

\subsection{Analysis and Discussion}
\subsubsection*{Discussion on the Variance of Methods}
Our investigation into the stability of the methods above revealed that the DSER method, in particular, exhibits significant performance variance. This inherent instability makes evaluation based on a single run unreliable. To obtain a more robust and consistent measure of performance, we aggregated results across multiple experiments. Therefore, the accuracy for the DSER method is reported as the average of 16 independent DSER runs in our main experiment.

\begin{figure}[H]
    \centering
    \includegraphics[width=1\linewidth]{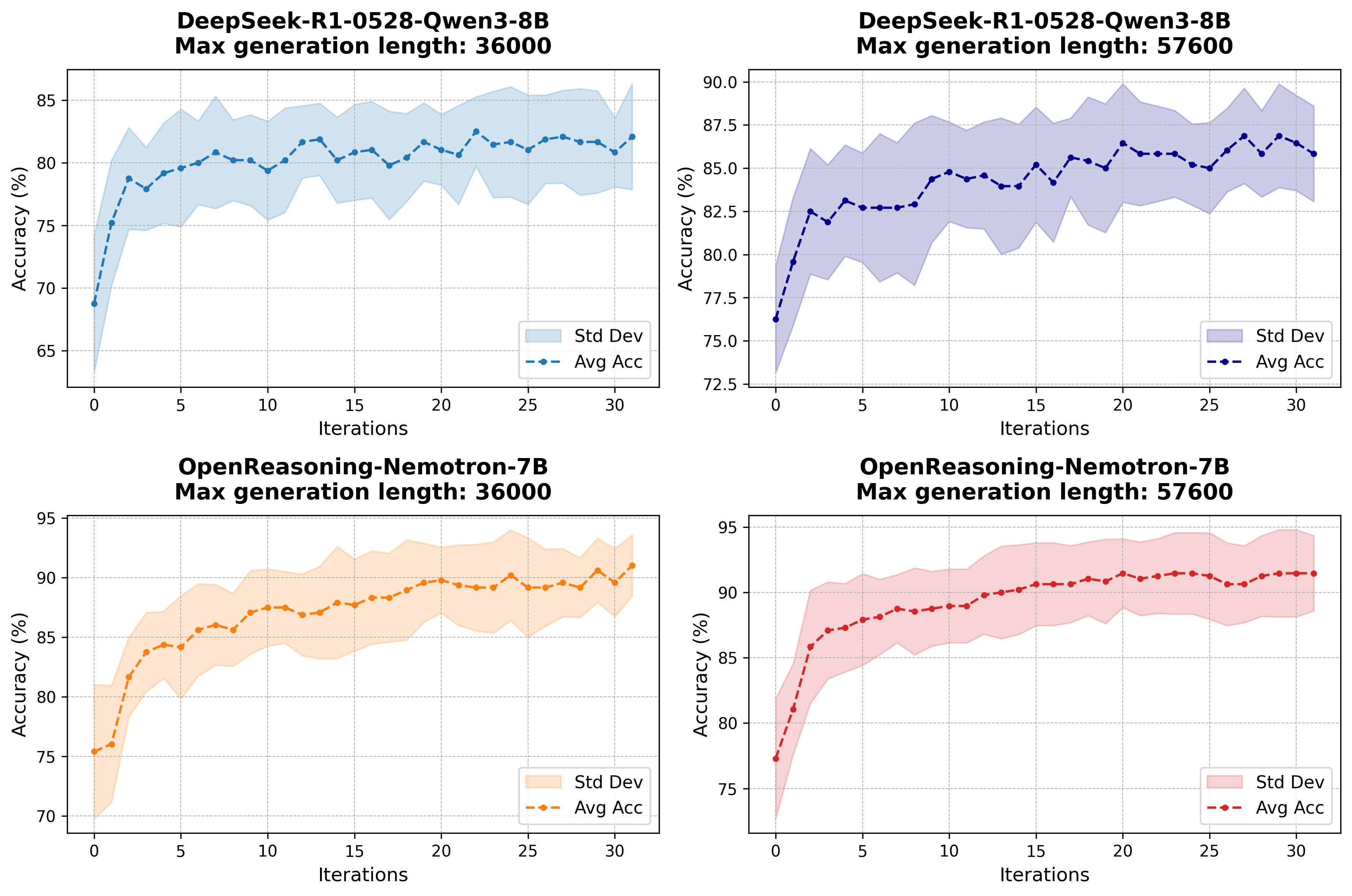}
    \caption{\textbf{Performance Evolution and Instability Analysis of the DSER Method.} The figure illustrates the performance curves of the DSER method evolving on the AIME25 dataset with two different LLMs. The solid line represents the mean accuracy over 16 independent trials, and its upward trend with more iterations demonstrates the method's effectiveness. However, the wide shaded area (mean ± one standard deviation) reveals significant performance variance and instability across different runs.}
    \label{fig:dser_var}
\end{figure}

We conducted experiments on the AIME25 dataset with the \path{OpenReasoning-Nemotron-7B} and \path{DeepSeek-R1-0528-Qwen3-8B} models, setting the maximum generation length to 36000 and 57600. The results, averaged over 16 trials, are plotted in Figure~\ref{fig:dser_var}. In the figures, the solid line represents the mean accuracy, which consistently rises until convergence with the number of iterations despite fluctuating, demonstrating the effectiveness of the self-evolving process. However, the shaded area, indicating the range from mean minus to mean plus one standard deviation of 16 independent trials, is notably wide across all configurations. This large variance underscores a lack of stability in the method's performance. The detail numbers can be seen in Table~\ref{tab:dser_var}.

This observed instability is a reasonable and inherent characteristic of the DSER framework.  Then the core of the method involves querying an LLM to verify and correct its own solutions, which introduces substantial stochasticity. LLMs have inherent limitations in their reasoning abilities; furthermore, their bias towards their own answers may cause them to overlook their own mistakes~\citep{deepseek-math-v2}. This inherent stochasticity in the self-evolving loop is the primary source of the high performance variance observed in the results.

\begin{table}[h]
    \centering
    \caption{\textbf{Stability of DSER Performance.} The table presents the mean accuracy and standard deviation for two LLMs evaluated at two different context lengths. All results are reported after 32 iterations.}

    \label{tab:dser_var}
    \renewcommand{\arraystretch}{1.2} 
    \begin{tabular}{llcc}
        \toprule
        \textbf{Model} & \textbf{Length } & \textbf{Mean Acc (\%)} & \textbf{Std Dev (\%)} \\
        \midrule
        \multirow{2}{*}{DeepSeek-R1-Qwen3-8B} 
          & 36,000 & 82.08 & 4.23 \\ 
          & 57,600 & 85.83 & 2.76 \\
        \midrule
        \multirow{2}{*}{OpenReasoning-Nemotron-7B} 
          & 36,000 & 91.04 & 2.56 \\
          & 57,600 & 91.46 & 2.88 \\
        \bottomrule
    \end{tabular}
\end{table}





\begin{figure}[H]
    \centering
    \includegraphics[width=0.85\linewidth]{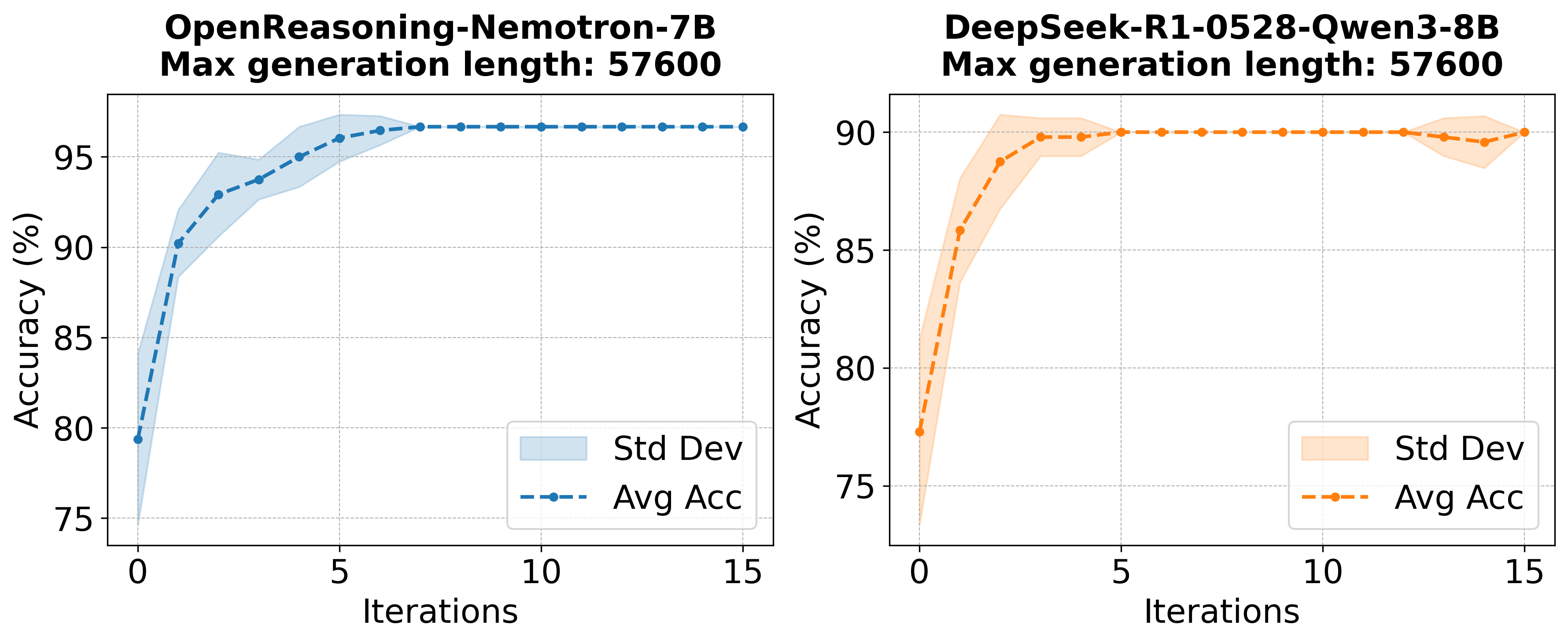}
    \caption{\textbf{Performance Evolution and Instability Analysis of the Population-Evolve Method.}}
    \label{fig:var_PE}
\end{figure}

\subsubsection*{Comparison of Population-Evolve and DSER on Stability and Efficiency}
Our method exhibits superior convergence speed and stability over DSER. This advantage comes from the eventual consensus reached among the solutions in the population, which leads to a dramatic drop in variance upon stabilization (evident in Figure~\ref{fig:var_PE}).

At the same time, we find that DSER method are extremely time-consuming. Specifically, each iteration in DSER consists of two distinct stages: a self-verification stage and a self-correction stage. These stages are executed sequentially, which makes each iteration substantially time-consuming. 

While our proposed method, Population Evolve, offers a notable advantage in terms of time efficiency. This is primarily because our approach requires only generating operation per iteration. 
Furthermore, it exhibits a rapid convergence rate, typically reaching convergence within just 4 to 8 rounds. 

\begin{figure}[H]
    \centering
    \includegraphics[width=0.83\linewidth]{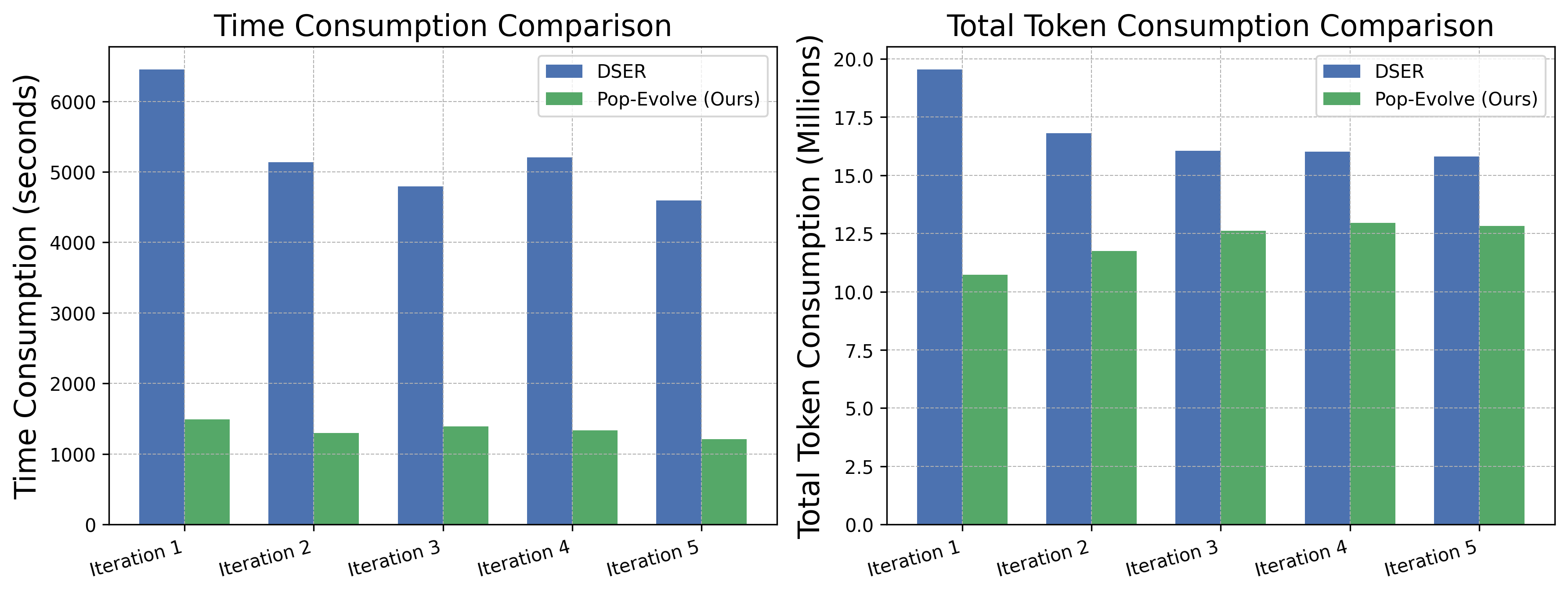}
    \caption{\textbf{Efficiency Comparison between the Proposed Method and DSER.} This figure illustrates the comparison of total runtime and token consumption (input and output) between our method and DSER. The analysis excludes the initial generation phase (iteration 0), which is common to both methods, to highlight the efficiency gains in the subsequent iterative process.}
    \label{fig:token_time}
\end{figure}

Figure \ref{fig:token_time} provides evidence for the efficiency claim by comparing the runtime and total token consumption against the DSER baseline in each iteration, clearly demonstrating the efficiency gains of our approach, an average time reduction of $74.13\%$.


\subsection{Ablation Study}


\begin{figure}[H]
    \centering
    \includegraphics[width=0.8\linewidth]{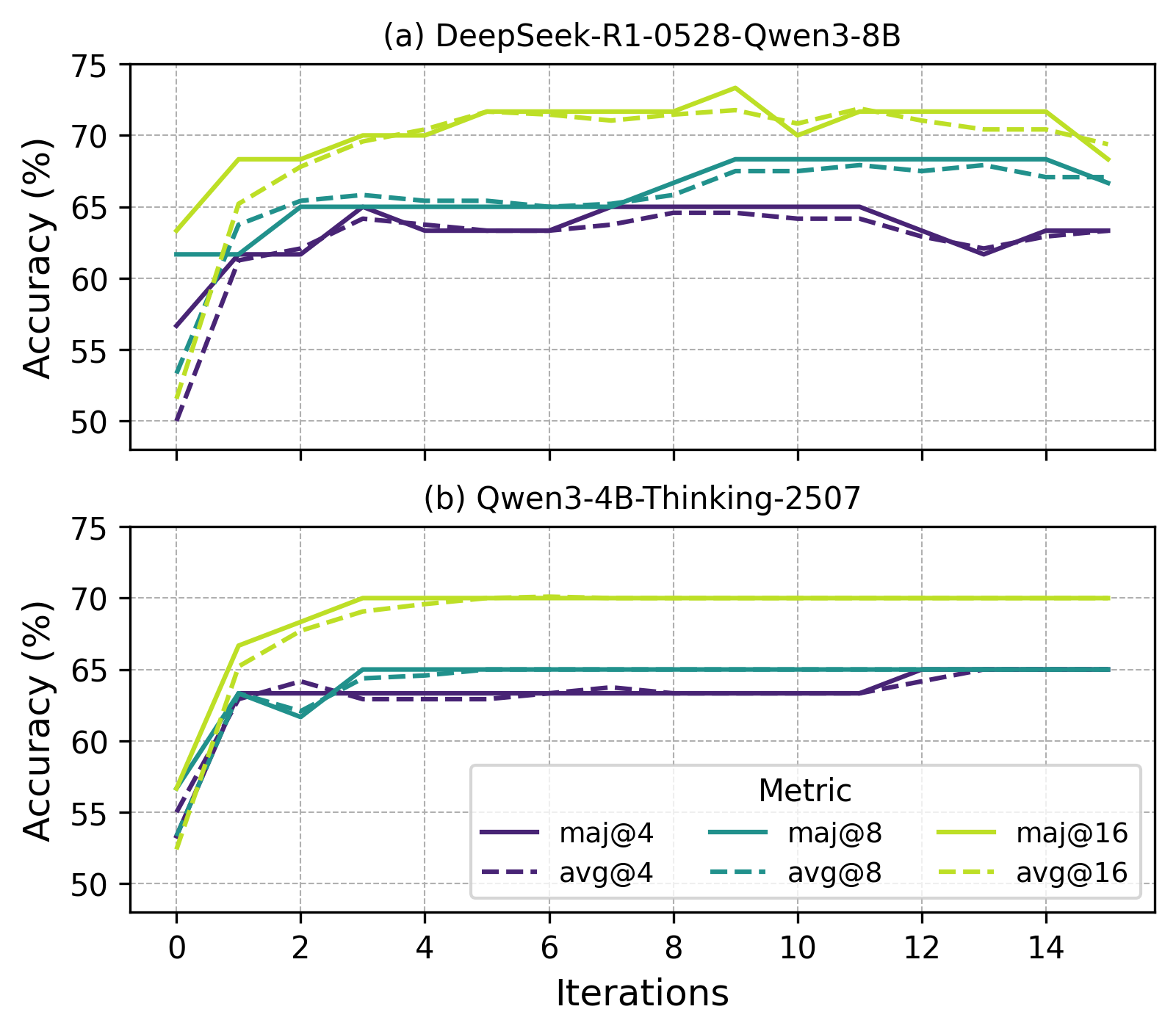}
    \caption{\textbf{Ablation Study on Population Size (P)}}
    \label{fig:ablation_p}
\end{figure}

We conducted an ablation study on the population size, $P$, to investigate its impact on our method's performance. Intuitively, a larger population size enhances the diversity of solutions, thus increasing the likelihood of including the correct solution. This, in turn, enables the LLM to generate a higher-quality subsequent generation based on the previous one. 

Our experimental setup involved the HMMT24-25 dataset, with two models: \path{Qwen3-4B-Thinking-2507} and \path{DeepSeek-R1-0528-Distill-Qwen3-8B}. We tested population sizes of P = 4, 8, and 16 on one trail. As shown in the figure~\ref{fig:ablation_p}, we plotted the accuracy against the number of iteration rounds.

The experimental results are shown in Figure~\ref{fig:ablation_p}. We observe a clear and consistent trend: as the population size $P$ increases, the performance of our method improves. This validates our hypothesis that a larger population enhances the diversity of solutions, thereby increasing the final accuracy.


\section{Conclusion}

In this work, we present Population-Evolve, a novel parallel sampling and evolutionary method that effectively harnesses the inherent comparative verification and reasoning abilities of Large Language Models.

By unifying several popular test-time scaling methods from the perspective of genetic algorithms, our framework offers a clear view and highlights the potential of test-time scaling for LLM reasoning tasks.

Empirical results on competition-level mathematical benchmarks demonstrate that our method significantly outperforms the naive majority voting baseline, consistently achieving superior performance on both the BRUMO, AIME and HMMT datasets. Moreover, our findings validate the robustness and generalizability of our approach across various models. 

In future work, we intend to explore end-to-end Reinforcement Learning to train LLMs specifically optimized for the Population-Evolve paradigm. Additionally, we aim to incorporate inference-time communication mechanisms—similar to Gemini's DeepThink—to facilitating higher-quality population evolution.
 

\bibliographystyle{abbrvnat}
\bibliography{references}

\appendix
\section{Appendix}

\subsection{Other Related Prompts}\label{ax:prompts}

\subsubsection{DSER related prompts}

\begin{promptbox}{Generation Prompt}
Please think step by step and put your final answer in \verb|\\boxed{{}}|.
\\\\
Question:
\verb|{question}|
\end{promptbox}

\begin{promptbox}{Verification Prompt}
Verify the given solution step by step to check correctness.
Provide a short verification report, containing the key points of the solution and any errors found. 

Finally, put your judgement strictly in the format: \verb|\\boxed{{correct}}| if correct, or \verb|\\boxed{{incorrect}}| if incorrect.
\\\\
Solution:
\verb|{solution}|
\end{promptbox}

\begin{promptbox}{Correction Prompt}
Given your previous solution and a verification report, reconsider the problem carefully and provide a corrected solution.
Output your final answer strictly in the format: \verb|\\boxed{{}}|.
\\\\
Previous solution:
\verb|{solution}|
\\\\
Verification report:
\verb|{verification_report}|
\end{promptbox}

\subsubsection{GenSelect related prompts}

\begin{promptbox}{Generation Prompt}
Please think step by step and put your final answer in \verb|\\boxed{{}}|.
\\\\
Question:
\verb|{question}|
\end{promptbox}

\begin{promptbox}{Selection Prompt}

You will be given a challenging math problem followed by \verb|{len(solutions)}| solutions. Your task is to systematically analyze these solutions to identify the most mathematically sound approach.
\\\\
Problem: \verb|{problem}|
\\\\
Solutions:
\verb|{solution_texts.strip()}|
\\
Evaluation Process:\\
1. Initial Screening: Group solutions by their final answers and identify mathematical contradictions.\\
2. Detailed Analysis: For remaining solutions, evaluate mathematical precision, logical progression, and completeness.\\
3. Solution Comparison: Compare viable solutions based on efficiency, clarity, and sophistication.
\\\\
Your response should include:\\
- Brief analysis of conflicting answers.\\
- Detailed evaluation of mathematically sound solutions.\\
- Justification for eliminating incorrect solutions.\\
- Clear explanation for selecting the best approach.\\
\\\\
End your evaluation with exactly:\\
Judgment: [IDX]
\\
where IDX is the index from 0 to \verb|{max_idx}| of the best solution.
\end{promptbox}



\end{document}